\documentclass[doc, dvipsnames]{apa7}

\usepackage[T1]{fontenc}
\sffamily

\usepackage{tabularx}
\newcolumntype{Y}{>{\centering\arraybackslash}X}
\usepackage{ragged2e}
\usepackage[american]{babel}

\usepackage{csquotes}
\usepackage{hyperref}
\hypersetup{
    colorlinks=true,
    linkcolor=blue,
    citecolor=teal, 
    filecolor=green,
    urlcolor=blue}

\usepackage[style=apa,sortcites=true,sorting=nyt,backend=biber, hyperref=true]{biblatex}

\DeclareLanguageMapping{american}{american-apa}

\title{Large Language Models in Numberland: A Quick Test of Their Numerical Reasoning Abilities}

\shorttitle{LLMs in Numberland}

\authorsnames[]{Roussel Rahman}
\authorsaffiliations{{Stanford University, Stanford, CA, USA}}

\leftheader{Rahman}

\abstract{
An essential element of human mathematical reasoning is our number sense -- an abstract understanding of numbers and their relationships -- which allows us to solve problems involving vast number spaces using limited computational resources. Mathematical reasoning of Large Language Models (LLMs) is often tested on high-level problems (such as Olympiad challenges, geometry, word problems, and puzzles), but their low-level number sense remains less explored. We introduce "Numberland," a 100-problem test to evaluate the numerical reasoning abilities of LLM-based agents. The tasks -- basic operations, advanced calculations (e.g., exponentiation, complex numbers), prime number checks, and the 24 game -- aim to test elementary skills and their integration in solving complex and uncertain problems. We evaluated five LLM-based agents: OpenAI’s o1 and o1-mini, Google Gemini, Microsoft Copilot, and Anthropic Claude. They scored 74–95\% on the first three tasks that allow deterministic steps to solutions. In the 24 game, which needs trial-and-error search, performance dropped to 10–73\%. We tested the top 24 solver (o1 with 73\% accuracy) on 25 harder problems, and its score fell to 27\%, confirming search as a bottleneck. These results, along with the types of mistakes, suggest a fragile number of LLMs, which is a bit surprising given their prowess in challenging benchmarks. The limits of LLM numerical reasoning highlight the scope of simple, targeted tests to evaluate and explain LLM math skills to ensure safe use.
}


\authornote{
   \addORCIDlink{Roussel Rahman}{0000-0003-2009-4246}

  Correspondence concerning this article should be addressed to Roussel Rahman, Department of Photon Science, SLAC National Accelerator Laboratory, Menlo Park, CA.  E-mail: \href{mailto: roussel.rahman@gmail.com}{roussel.rahman@gmail.com}}

\begin{document}
\maketitle

LLM-based agents 
have exceeded our expectations in performing creative tasks once considered uniquely human, such as composing essays, writing code, and solving mathematical problems. As they quickly become our assistants across devices and interfaces, understanding their limits is crucial for safe and efficient use. For example, when using LLM assistance in solving mathematical problems, we need to know how good they are as calculators. Can they add 2 and 2? Can they divide and multiply? How about finding the prime factors of a large number (e.g., $9999999999$)? Often, these agents remain oblivious to their limits or mistakes, making it essential for users to understand what they can or cannot do.

An early limit LLMs faced was mathematical reasoning, a skill studied extensively and improved greatly since then \parencite{Garisto2022}. However, a fundamental aspect of mathematical reasoning remains less explored: their numerical abilities. Number spaces are different from word spaces (the usual domain of LLMs) as they are essentially infinite in size, and the relationships between numbers are distinct from the relationships in text embedding spaces. Humans possess and develop an intriguing "number sense" -- an abstract understanding of what numbers represent, the relationships among them, and how their regularities can be exploited -- that allows us to solve complex problems efficiently \parencite{dehaene2001precis, dehaene2011number, berch2005making}.  
Whether LLMs possess a sense of numbers remains unexplored. However, a granular understanding of their low-level numerical reasoning that drives higher-level mathematical reasoning can not only help us estimate their limits, but also inform promising directions in explaining AI behavior, such as Mechanistic Interpretability (MI), to decode AI reasoning from its building blocks. 

In this work, we examine the number sense of five LLM-based agents in "Numberland" -- a test consisting of 100 problems that range from extremely simple (e.g., mathematical operations) to relatively complex (checking primality of numbers), and some require an informed trial-and-error search for solutions (the 24 game). Analyzing each agent's performance at multiple levels, we evaluate their basic numerical skills and their ability to combine and generalize the basic skills to solve increasingly complex problems. We find that, while the agents perform reasonably well in three of the four tasks, even advanced agents struggle in the trial-and-error search required in the 24 game, indicating a fragile number sense underneath their deterministic prowess.

The organization of the paper is as follows. We begin by reviewing research on LLM mathematical reasoning before moving to human mathematical reasoning and the role of number sense in it. After describing Numberland, the agents, and our methods, we closely examine LLM performance at different levels of granularity (e.g., total score, category scores, and type of mistakes). Based on the results, we highlight the limits of LLMs' number sense and the future directions to test, explain, and improve their mathematical reasoning.

\section{Mathematical Reasoning of LLMs and AIs}


Reasoning generally refers to the process of making a sequence of rational\footnote{Although defining rationality itself remains challenging, please see \textcite{gigerenzer2020bounded} for a discussion.} decisions to achieve some goals using available information and computational resources. Mathematical reasoning refers to the process of solving mathematical problems by efficiently using various mathematical concepts (e.g., representations, facts, and principles) and their relationships. We divide our review of LLM mathematical reasoning into four perspectives, starting at a high level and progressing to lower levels.

\subsection{Using Benchmark Scores}
Most commonly, LLM agents' mathematical abilities are evaluated by their average scores on benchmark datasets \parencite{ahn2024large}. These benchmarks include diverse problem types -- such as arithmetic, algebra, geometry, puzzles, and theorem proving -- of varying difficulty levels. For instance, the MATH dataset \parencite{hendrycks2021measuring} contains 12500 high-school math competition problems, while GSM8K \parencite{Cobbe2021} includes 8500 middle-school arithmetic problems. The Minif2f dataset \parencite{zheng2021minif2f} features 488 problems from prestigious mathematics contests, such as AIME, AMC, and IMO. Please see \textcite{ahn2024large} for an elaborate survey of the benchmarks used for evaluating LLM mathematical reasoning.

\subsection{Reasoning as a Component of General Intelligence}

\textcite{chollet2019measure} developed the Abstract Reasoning Corpus for Artificial General Intelligence (ARC AGI) test to measure general intelligence by the ability to transfer learning and solve new problems. On each problem, agents are shown several pixelated images (30x30 grids) in input-output pairs. The task is to learn the relationship in pairs and predict the outputs for new input images. Inspired by human IQ tests, it tests AGIs on a set of Core Knowledge \parencite{spelke2007core} that humans naturally possess (such as goal-directedness and an understanding of natural numbers) and combine with prior experience to solve problems.


While the ARC AGI and other benchmarks proved difficult for the then State-of-the-Art (SOTA) agents, LLMs have improved rapidly on these benchmarks \parencite{Garisto2022, chollet2024arc}. A large set of reasoning benchmarks with updated lists of top performers can be found here: \url{https://paperswithcode.com/area/reasoning}. 
Importantly, aggregating performance over a large number of problems makes it difficult to pinpoint the source of improvements \parencite{rahman2021precisemeasures, rahman_gray2020topics}, leading researchers to seek more granular explanations of AI reasoning.



\subsection{Using Chains and Trees of Reasoning with Reinforcement Learning}
\textcite{wei2022chain} showed \textit{Chain of Thought} (CoT) prompting -- asking models to divide complex problems into smaller elements and reason step by step -- considerably improves reasoning performance without additional training. This approach combines a divide-and-conquer strategy with think-aloud protocols (often used in studying human reasoning), and the self-reported steps serve as explanations of reasoning. Extending CoT, \textcite{yao2023tree} introduced the \textit{Tree of Thoughts} (ToT) approach that enables goal-directed problem solving in LLMs by representing problems as trees of sub-problems before systematically searching through multiple reasoning paths. Recent efforts build on the tree-search perspective using reinforcement learning from human feedback \parencite[e.g., ][]{bai2022training} or through self-reflections \parencite[e.g.,][]{ouyang2022training}; please see \textcite{plaat2024reasoning} for a review. Both CoT and ToT reasoning, along with reinforcement learning, have become integral elements of modern LLM agents' reasoning. Notably, the ToT approach is built upon pioneering works by \parencite{newell1958elements, newell1959report} on complex problem solving (Section \ref{sec:cps}), which also forms the basis of our explorations.

\subsection{Using Mechanistic Interpretations}

Beyond LLM self-reports, Explainable AI (XAI) offers several classes of tools of varying granularity to investigate reasoning \parencite{bereska2024mechanistic}. A particularly promising approach has been MI, which seeks to form mechanistic interpretations of behavior by reverse-engineering MLP-based AI models from neuron-level information. MI focuses on identifying building blocks -- such as disentangling \textit{features} from neurons, \textit{circuits} or weighted subgraphs of features, and \textit{motifs} or recurring patterns composed of circuits and features -- and reconstruct behavior bottom-up as means to develop human-interpretable explanations \parencite{olah2020zoom, nanda2023progress, bereska2024mechanistic}. 
\textcite{olsson2022context} provided a circuit-level explanation of pattern recognition, identifying specific \textit{induction heads} in transformers that enable in-context learning. 
\textcite{bricken2023towards} decomposed neural activations into interpretable features corresponding to distinct reasoning operations. 
\textcite{li2023emergent} demonstrated that models develop structured internal representations of complex problem environments encoding spatial relationships and causal dynamics. \textcite{stolfo2023mechanistic} explored the causal roots of arithmetic reasoning by intervening in the activations of different neurons and examining the changes in prediction probabilities. \textcite{nanda2023progress} reverse-engineered an algorithm learned by a single-layer transformer for modular addition (i.e., problems of the form $mod(a+b, n)$). MI approaches also provide insights into LLM limits. For example, the model tested by \textcite{nanda2023progress} learned an algorithm for modular addition --
involving trigonometric identities with discrete Fourier transforms -- much more complex than how humans may solve it. 
\textcite{dziri2023faith} highlights that transformer models often rely on pattern matching rather than true compositional reasoning, revealing the boundaries of their reasoning capabilities.

\section{Goals and Scope of Contributions}

\subsection*{Scope: Targeting Persisting Traits of Reasoning for Integrated Explanations}

Evidently, explanations of LLM reasoning have been diverse, each with unique advantages and challenges. While CoT and ToT produce readable reasoning traces, the true reasoning processes may diverge from agent self-reports \parencite{ahn2024large, turpin2023language}. In contrast, MI offers specificity but faces scalability challenges, as current methods lack automation and require extensive manual effort. Consequently, while simple behavior can be comprehensively explained, complex behavior is often explained in broad strokes \parencite{wang2022interpretability, bereska2024mechanistic, bricken2023towards}. To scale up MI, multi-level analysis can aid in using top-down, macroscopic observations to guide the microscopic, bottom-up views of MI, increasing efficiency \parencite{bereska2024mechanistic}. For instance, if different agents make similar mistakes in a domain (such as mathematical reasoning), their causal roots may be traced through different levels, such as high-level performance, self-explanations, and low-level elements. Identifying such general or persistent characteristics through targeted experiments can guide our search for the building blocks and help progress towards an integrated understanding.


\subsection*{Goal 1: A foundation for Multi-Level Investigations of LLM (Mathematical) Reasoning}

We adopt a "divide-and-reconstruct" approach inspired by studies of human problem-solving and mechanistic interpretations of AI reasoning -- breaking complex tasks into elementary skills before evaluating their individual use and their integration in complex skills. As we will discuss next, search in complex, uncertain environments is challenging for both artificial and human agents. As problem complexity prohibits exhaustive search, reasoning can be considered as the ability to efficiently navigate this complex search space within computational limits \parencite{gigerenzer2020bounded, chollet2019measure, lieder2020resource, simon1971human, yao2023tree}.

\subsection*{Goal 2: Test the limits of LLM Numerical Reasoning Abilities or "Number Sense"} 

Specific to mathematical reasoning, problems involving trial-and-error search (e.g., Game of 24, cryptoarithmetic, and Sudoku) remain a challenge for LLMs. On simple to relatively hard games of 24, CoT scored only 3\%, whereas ToT scored 74\% \parencite{yao2023tree}. As we will show, even the SOTA LLM-based agents falter in sufficiently hard 24 problems that are still simple for humans due to their number sense. To trace the causal root, we designed "Numberland" to test skills in fundamental mathematical operations and their integration in two complex problems -- (a) check primality of numbers, a task with deterministic algorithms to search for solutions but non-deterministic endpoints, and (b) the 24 game. This way, we elucidate strengths and weakpoints in LLMs' elementary numerical skills that contribute to high-level performance.




\section{Human Mathematical Reasoning and Problem Solving}

\subsection{Number Sense and Mathematical Reasoning}
Number sense refers to an intuitive understanding of numbers, their relationships, and how they can be manipulated. Functionally, it allows us to quickly estimate, compare, and reason about numbers without formal or exact calculations \parencite{dehaene2001precis}. The origins of human number sense have been explored from different perspectives. For example, \textcite{gallistel1992preverbal} described two parallel systems for numerical processing: a preverbal system to approximate quantities that emerges early in development and a verbal system for precise calculations that develops with cultural learning. \textcite{dehaene1992varieties} proposed a triple code model for number representations in three forms: (1) visual with numerals (e.g., "37"), (2) verbal ("thirty-seven"), and (3) analogical representations as distributions of activations along the number line. For alternate but overlapping perspectives, please see \textcite{spelke2007core, butterworth2005development, karmiloff1994precis, wynn1995origins}. 

While our number sense is abstract, neuroimaging studies (using positron tomography and fMRI) reveal a specific brain region -- the inferior parietal cortex -- that contributes heavily to it \parencite{dehaene1996organization, kiefer1997time, pinel1999event}. This region activates when humans calculate \parencite{roland1985localization, chochon1999differential, dehaene1999sources, pinel1999event, rueckert1996visualizing} or even represent numbers in mind \parencite{dehaene1996organization, kiefer1997time, pinel1999event}. This brain region is not the only one contributing to numerical or mathematical reasoning; rather, multiple brain regions combine to make up our complex abilities \parencite{dehaene2001precis}.

Important takeaways are that a part of our number sense is biologically determined and has evolutionary roots, and a part of it is learned and developed. 
Notably, a common theme is that the studies of human reasoning and developmental changes sought process-level explanations, as opposed to a black box relating input abilities and output performance. For example, children learning addition have been observed to adaptively use at least 5 different strategies, and their frequency of using these strategies changes with learning \parencite{siegler1987perils, siegler1989children}. The adaptive use of these strategies and their transitions with learning has been emulated in cognitive models \parencite{shrager1998scads}. However, similar to the challenges to MI for LLMs, scaling up the explanations for more complex problems remains challenging. Rather, complex problem solving has been the focus in studies of one particular type of human rationality: Bounded rationality.

\subsection{Bounded Rationality and Solving Complex Problems}\label{sec:cps}
Complex problems generally consist of many interconnected sub-problems in a hierarchical network, in which the high-level sub-problems serve as goals for immediately lower-level ones \parencite{simon1962architecture}. Consequently, there are numerous paths from the lowest-level components to the highest-level goals. Complex problem solving can be described as a search for paths in the hierarchy based on subgoals \parencite{newell1958elements, newell1959report, simon1971human}. A key point is that it is not possible for humans -- or any computationally limited agents -- to perfectly reason their way to optimal paths in all complex search spaces \parencite{savage1954foundation, simon1975optimal, gigerenzer2008heuristics}. When optimality is beyond reach, humans demonstrate rationality bounded by their computational limits -- namely, bounded rationality \parencite{simon1955behavioral, gigerenzer2002bounded, gigerenzer2020bounded}. Specifically, they solve complex problems efficiently using a set of heuristics to approximate solutions that "satisfice" (i.e., satisfy and suffice) for their needs \parencite{simon1955behavioral, simon1971human, newell1958elements, gigerenzer2001adaptive, gigerenzer2020bounded}. From this perspective, human complex skills are built using many heuristics as elements to tackle different parts of complex problems. The satisficing process can be imagined as an iterative search in the hierarchical tree for satisfactory solutions using heuristics as guides \parencite{gigerenzer2008heuristics}.  

A vast amount of evidence shows that the boundedly rational humans can solve complex problems accurately and efficiently despite their obvious computational limits and frequently outperform the SOTA algorithms tailored for the problems \parencite{gigerenzer2008heuristics, bossaerts2017computational, rahman2022dynamics}. In contrast, investigations of LLM reasoning in complex problems (e.g., Graph Coloring, Cryptoarithmetic, Sudoku, and the Game of 24) show struggles with complexity and a disparity between human and LLM performance without fine-tuning for specific tasks \parencite{mittal2024puzzlebench, giadikiaroglou2024puzzle, ding2023everything}. 


Understanding how LLMs behave and reason in complex problems through their building blocks -- as done for humans -- can help us explore ways to improve their reasoning and bridge this gap. A frequent observation from studies of human problem-solving is that a complete explanation of behavior needs to specify the processes that generate the behavior \parencite[]{newell1958elements, simon1971human, anzai1979theory, siegler1987perils, siegler1991microgenetic, agre90csc, gigerenzer2020explain}.
This requirement again highlights the promise of XAI tools -- especially, MI -- in explaining (mathematical) reasoning of LLMs. Observations from experimental studies of human reasoning are validated by simulating behavior in cognitive models \parencite{sun2008introduction}. Similarly, circuit-based MI investigations provide a way to validate observations by reverse engineering LLM behavior in explanatory models.


\section{Methods}

\subsection{Experimental Task: The "Numberland"}

The Numberland contains 100 problems divided equally into four sets by problem type, with a goal to examine elementary mathematical skills in the first two sets before examining their combined use in solving more complex problems. In each set, the problems are indexed approximately by problem difficulty as a sub-level of analysis.


\subparagraph{Set 1: Basic Mathematical operations}
The problems involve only the four basic mathematical operations: addition, subtraction, multiplication, and division. We begin with simple problems (e.g., 2 + 2 + 4 - 2 = ? and 1/13 = ?) before making the problems slightly more difficult by involving larger sets of numbers and transcendental numbers (such as $\pi$ and $e$). Some problems include reciprocals of prime numbers to exploit a numerical precision problem often observed in LLMs.

\subparagraph{Set 2: Advanced Mathematical Operations}
We introduce more advanced operations and concepts -- specifically, exponentiation, logarithms, and complex numbers. Most problems require combinations of operations. In some problems, slightly unusual values, such as fractions as exponents (e.g., $400^{0.23}$) and logarithms with base 5 or 13 (e.g., $\log_{13}{169}$).

\subparagraph{Set 3: Check if a Number is Prime}
Here, we ask the agents to check the primality of numbers. We varied the length of the numbers (discontinuously from single digits to 54-digit numbers). Some numbers are relatively famous and likely to be discussed in LLM training data, such as Mersenne Primes of the form $2^p-1$ (such as $2^{13}-1$ and $2^{31} - 1$). We also include some non-primes of the same form ($2^{27}-1$ and $2^{29}-1$), which we refer to as the \textit{Mersenne Prime Lookalikes}. Some long primes were copied from the Wikipedia page on prime numbers, while some prime and non-prime numbers were hand-picked as randomly as we could.

Why use prime-checking? This set allows us to examine LLM performance in combining basic skills to solve a more complex problem. The algorithms for checking the primality of numbers by searching for factors are well-known, but the point to stop searching is non-deterministic. There are explicit clues in (prime) number theory to exploit -- for example, any number of the form $2^p-1$ cannot be prime if $p$ is not prime (i.e., $2^{27} -1$ is not a prime) -- or more implicit ones in our number sense (e.g., we can tell 333 or 77777 are not prime without explicit reasoning).

\subparagraph{Set 4: The 24 Game}

$24$ is a popular number game that has been previously used in teaching children and testing LLMs \parencite{van2017cognitive, ding2023everything}. The game is simple and requires only elementary mathematical knowledge to play. A player is given a set of four numbers (e.g., $[2, 4, 6, 6]$ and their goal is to produce 24 as the output, using each number once, and the four basic operations -- addition, subtraction, multiplication, and division -- as many times needed (e.g., $(6+2-4) \times6 = 24$). 

Why use the 24 game? Despite its simplicity, the search space is quite vast, with numerous combinations of numbers and operations to consider. This space can be reduced using number relationships as clues. For example, a helpful strategy is to look for factors that make up $24$ or for multiples of $24$. In the above example, we may set aside one of the $6s$ from the set and try to manipulate the rest of the numbers to get the required $24/6=4$.
There are numerous such clues to exploit in the number system in developing strategies. However, none of the strategies guarantee success; rather, solvers need to find the solutions through an informed trial-and-error search using their "number sense."

\subsection{Participants: Five LLM-based Agents}

All tests were conducted over a 7-day period starting from December 30, 2024, PST. We began with eight LLMs and performed an initial test of competence on basic mathematical operations for inclusion in the main test. Three models were excluded as they did not score beyond our threshold of 80\%: ChatGPT 4o, Cohere AI, and Mistral AI. The rest of the five models (listed in Table \ref{tab:model_details}) were included in the Numberland. In all cases, we used graphical web applications to communicate with the agents. Except for the ChatGPT o1 models, we used the free versions of the models. To access the o1 models, we subscribed to ChatGPT Pro.

Note that the LLM agents differ substantially in architecture, and the differences further propagate due to fine-tuning for specific abilities. For example, the two o1 models are equipped with enhanced reasoning capabilities, while the o1 model "thinks" longer and uses more chains than the o1-mini model. The remaining three models we used are trained for general-purpose usage. Complete details are unavailable due to the proprietary nature of the models. Some high-level similarities are: (1) all use transformer models pre-trained on large corpora of texts, and (2) follow probabilistic principles to generate the next word using contextual information captured by positional encodings, self-attention mechanisms, and multi-head attention.

\begin{table}[!t]
\caption{Five LLMs included in the main test and their versions. All tests were conducted over a 7-day period starting from December 30, 2024, PST.}
\centering
\begin{tabularx}{\textwidth}{|Y|Y|Y|}
\hline
\textbf{Model}    & \textbf{Version}	& \textbf{Reasoning Enabled?}	\\ \hline
OpenAI ChatGPT    & o1-mini    			& Yes 				\\ \hline
OpenAI ChatGPT    & o1         			& Yes 				\\ \hline
Google Gemini     & 1.5        			& No 				\\ \hline
Anthropic Claude  & Sonnet 3.7 			& No 				\\ \hline
Microsoft Copilot & -          			& No 				\\ \hline
\end{tabularx}
\label{tab:model_details}
\end{table}

\subsection{Experimental Procedure and Scoring}

Each agent performed the Numberland test three times. We used six prompts in sequence to present the test to the agents (please see Appendix \ref{sec:prompts_used}). In the first prompt, we explained that the agents would be provided with a list of problems to solve and were asked to show their steps before summarizing their answers in an ordered list. In prompts 2-4, we presented the first three problem sets; in each prompt, we provided brief explanations of the task to perform and a set of 25 problems. In prompt 5, we asked the agents if they were familiar with the 24 game and its standard rules. After confirming they had the right version in mind, we gave them 25 games to play. We also conducted a follow-up test for our best-performing model on a set of 25 harder games of 24 (Appendix \ref{sec:prompts_used_followup}).

We score each correct answer by 1 point and a wrong answer by 0. For decimal fractions as answers, we asked the agents to provide three digits after the decimal point, but we gave them a point if the first two digits were correct. The scores reported here are averages over three trials.
Detailed results are provided in this repository: \url{https://osf.io/fncyu/}



\section{Results}

\subsection{Overall Test Performance of LLMs}

\begin{figure}[!t]
    \includegraphics[width=1.0\textwidth]
    {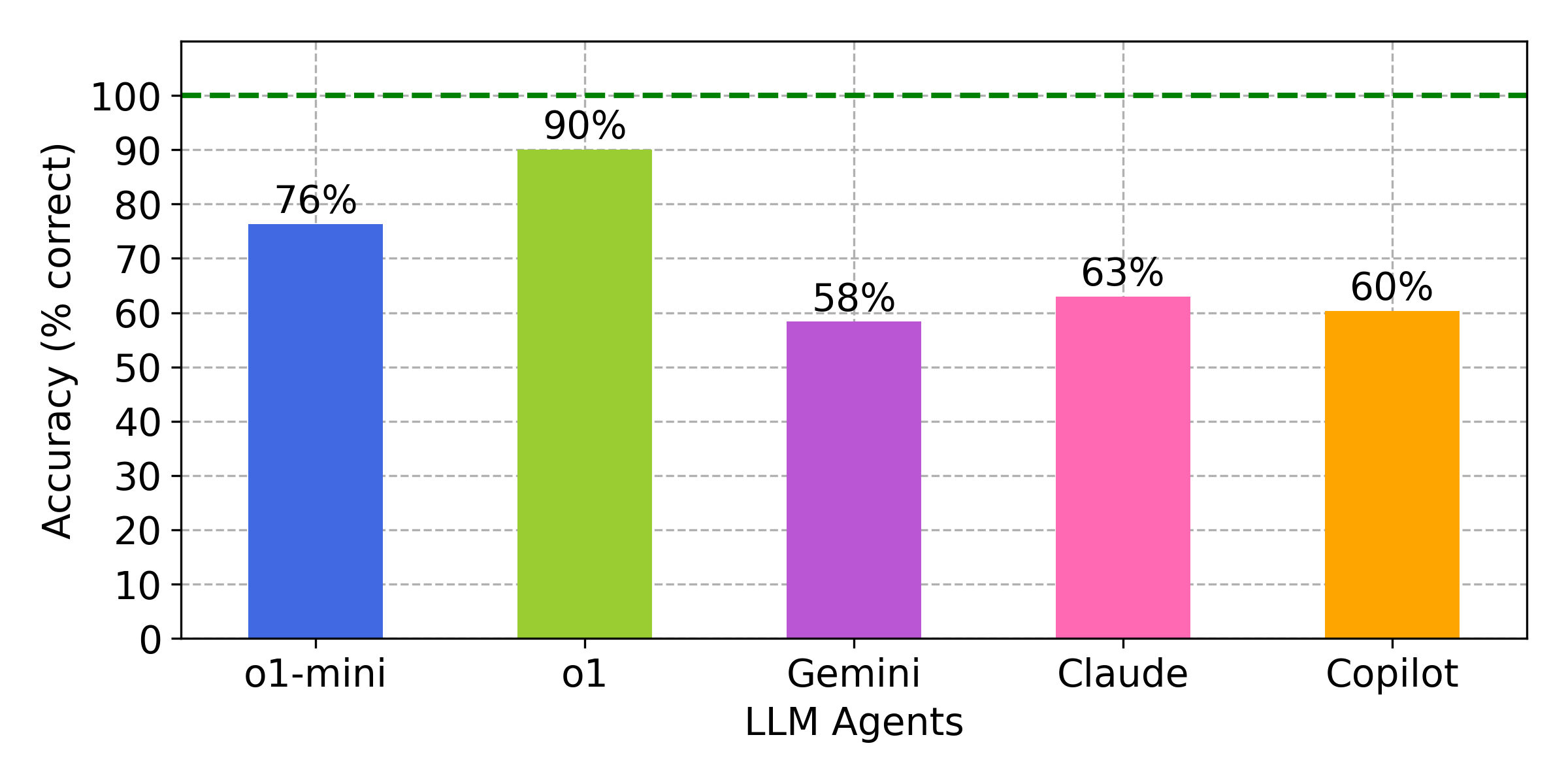}
    \caption{Overall performance of the models represented by the average score across problem sets.}
    \label{fig:overall_scores}
\end{figure}

The overall performance of the agents is shown in Figure \ref{fig:overall_scores}. ChatGPT o1 model performed the best (90\% average accuracy), followed by o1-mini (76\%). The remaining three models performed similarly to each other (58-63\%) but at a level considerably poorer than the two ChatGPT models. The performance discrepancy is not surprising since the ChatGPT models are equipped with explicit reasoning abilities. 
However, this high-level view obscures the types of mistakes made and the generality of mistakes across agents, details that are essential to estimate their competence in specific mathematical tasks. For example, we cannot tell which 10\% of the problems the o1 agent made mistakes, nor can we tell whether Claude, Gemini, and Copilot made mistakes in the same problems, despite similar average performance. In the following two sections, we focus on specifying the scope of their mistakes.

\subsection{LLM Performance in Different Problem Types}

\begin{figure}[!t]
    \includegraphics[width=1.0\textwidth]
    {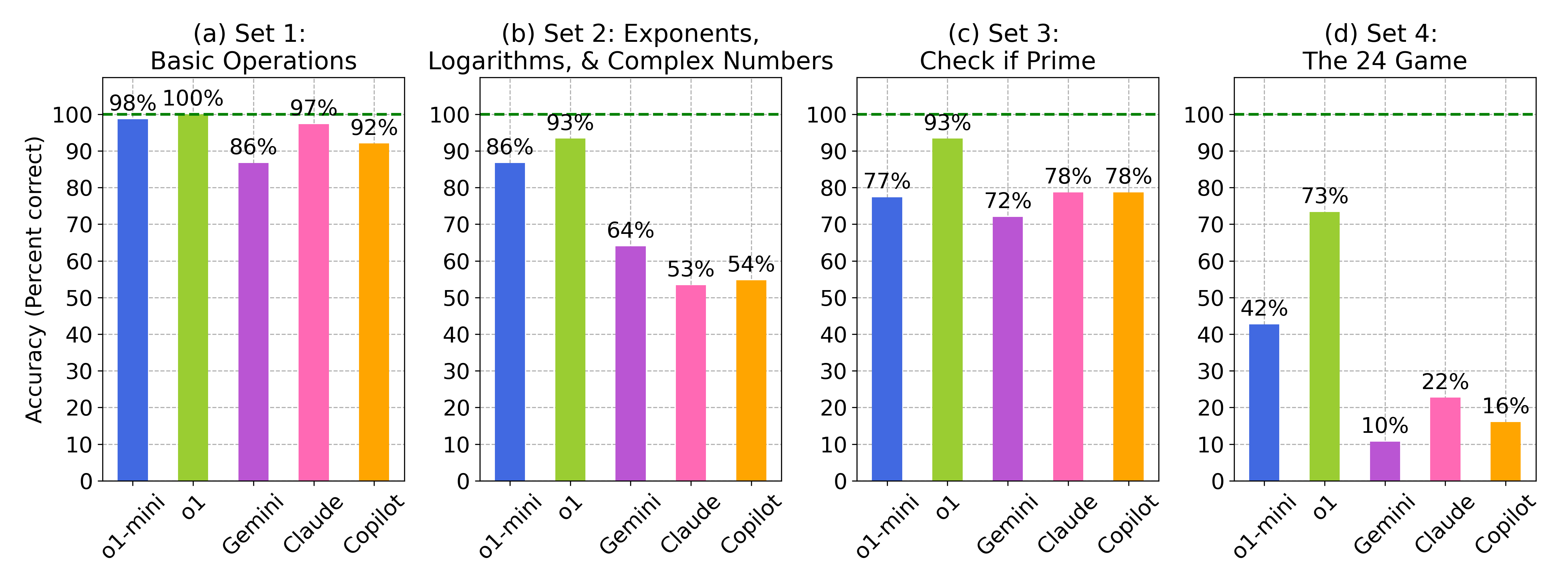}
    \caption{Performance of each agent in each of the four problem categories. The agents performed reasonably well in the first three sets but struggled in the fourth set containing the games of 24.}
    \label{fig:performance_by_category}
\end{figure}

Figure \ref{fig:performance_by_category} shows performance on different sets divided by problem type. All agents can be observed to perform the basic operations (Set 1) quite well (Fig. \ref{fig:performance_by_category}a), with the lowest accuracy being 86\%. As mentioned earlier, we used a cut-off of 80\% accuracy in this set as an inclusion criterion for the LLMs. This threshold ensures that all agents included in the test possess the elementary mathematical skills needed to perform the other tasks in the test. Therefore, the scores in the other tasks reflect how well they were able to generalize the elementary skills needed to solve more complex problems.

Set 2 (Fig. \ref{fig:performance_by_category}b) consists of more advanced problems involving exponents, logarithms, and complex numbers. On this set, the two ChatGPT models performed reasonably well, both scoring above 80\%, with only a slight drop from the first set. On the other hand, the scores of Gemini, Claude, and Copilot agents drop considerably. These three agents scored between 53-64\%, in stark contrast to their above 85\% average scores on the first set.

On Set 3 (Fig. \ref{fig:performance_by_category}c), we examined the agents' ability to use the knowledge of factors (tested in performing basic operations) to check if the given numbers are prime. As we can see, the o1 model retained its superior performance level on Set 3. On the other hand, the o1-mini agent dropped down to the level of the other three agents, who slightly raised their performance from the advanced operations set.

On Set 4 (Fig. \ref{fig:performance_by_category}d), we asked the agents to play 25 rounds of the 24 game. As we see, the performance of all LLM agents dropped sharply in this set compared to the first three sets. To provide a comparison, on average, all agents scored between 74\% and 95\% on the first three sets, whereas the agents scored between 11\% and 73\% in the games of 24. The o1 model is the top performer at 73\%, whereas all other agents solved less than 50\% of the problems.

These results show that while the agents perform reasonably well in fundamental operations and following deterministic steps to solutions (as reflected in the first three tasks), they struggle to transfer these skills to solve the games of 24. Notably, the 24 game does not require any new operations; rather, the basic operations tested on Set 1 suffice for this game. However, unlike the first three tasks, the 24 game also requires a trial-and-error search for solutions, which has been noted as a challenge for LLMs. In the next section, we examine the mistakes made in each problem type to specify the points of their struggles.


\subsection{A closer look at the types of mistakes made}

\begin{figure}[!t]
    \includegraphics[width=1.0\textwidth]
    {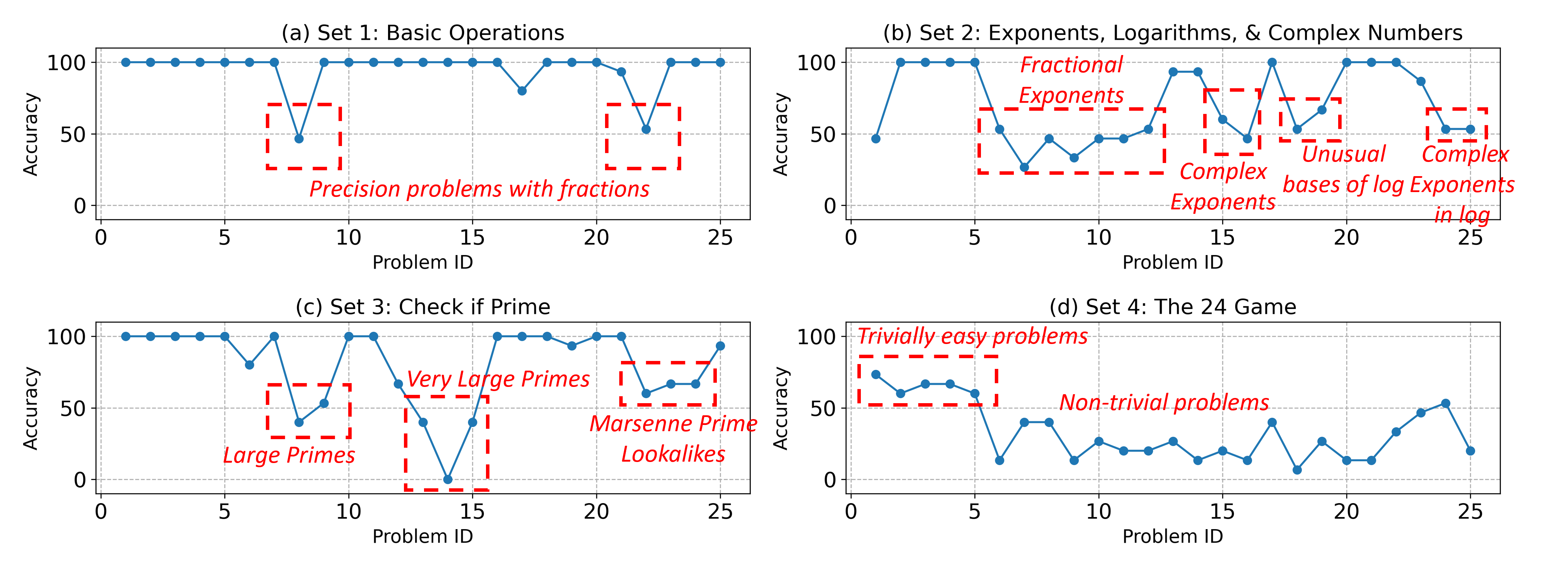}
    \caption{Average scores by problems in each category. As can be seen, some problems were harder than the rest for the agents. Annotations include problem features as the possible reasons for observed performance. Please see the main text for details.}
    \label{fig:mistakes}
\end{figure}

Average LLM performance in each problem is presented in Figure \ref{fig:mistakes}. For these results, we averaged the LLM performance over all agents. On Set 1 of basic operations, all LLMs can be observed to solve most problems perfectly. The mistakes were mainly limited to loss of precision in decimal fractions. These mistakes could be avoided by symbolic operations to simplify the problems, a strategy often adopted by the agents.

On the second set, the LLMs can be observed to struggle with multiple types of operations. An Achilles' heel lay in the fractional exponents (such as $400^{0.23}$), with which most LLMs -- barring the ChatGPT agents -- struggled. The agents also made more mistakes in problems involving unusual bases of logarithms and complex numbers.

In Set 3, the LLMs performed generally better than they did in Set 2, but again showed some common pitfalls. The most common mistake predictably coincided with large prime numbers. On the other hand, they were unaffected by the size of the numbers when dealing with Mersenne primes, but they often mistook the Mersenne Prime Lookalikes (i.e., the non-primes of the form $2^p-1$) for prime numbers.

On Set 4, which contains the games of 24, we see that LLM agents perform considerably worse than on the first three sets. Their performance can be observed to fall off a cliff after the initial few problems that were trivially easy and could be solved by simply multiplying the given numbers (such as [1,1,1, 24] or [1,1,3,8]). Surprisingly, some agents were not able to solve even these problems. The rest of the problems were more complex than that but still relatively easy for humans (presented in Appendix \ref{sec:prompts_used}), but all LLM agents showed less than perfect performance in these problems. A common mistake was assuming a solution did not exist for many of the problems, whereas all problems contained at least one solution. Moreover, the agents frequently broke the rules of the game to achieve 24, mainly in two ways: (1) not using all the numbers and (2) using some numbers more than once. Finally, another common set of mistakes was miscalculating expressions and then providing solutions that do not yield 24 without realizing the miscalculations. These mistakes are particularly surprising as they indicate a breakdown in basic mathematical skills, which we had observed the agents to possess when tested on simpler problems. Notably, the ChatGPT models did not make the last kind of mistakes and performed generally better in playing the 24 game.

\subsection{Follow-up Tests on a Harder Set of 24 Games}

To verify that the LLMs struggled with the trial-and-error search required in the 24 game, we tested our best-performing model -- ChatGPT o1 -- on a harder set of games five times (Please see Appendix \ref{sec:prompts_used_followup} for the prompts used). Whereas the o1 agent scored 73\% on the easier set, its score dropped down to 27\% on the harder set of problems. Similar to before, the agent did not make mistakes in executing expressions but failed more often to find a solution and mistakenly decided that it did not exist.

\subparagraph{Updates on March 20, 2025}

All results up to here are based on tests performed until the first week of January 2025. As LLMs are evolving fast, we include an update after a quick test of whether newer models are better at playing 24 or if it remains a persisting challenge.

We presented the second, harder set of 24 games to four reasoning-equipped models: ChatGPT o1, ChatGPT o3-mini, DeepSeek R1, and Grok 3. The performance does not appear to have improved considerably from the o1 model tested earlier in January. Once again, the most common failure turned out to be the conclusion that solutions do not exist for the problems, even though all problems have at least one valid solution. These mistakes were closely followed by using a number twice or not using all numbers. However, some reasoning models demonstrated a new, contrasting pitfall -- being stuck in seemingly infinite chains of reasoning before crashing or giving up\footnote{As the agents failed to complete the whole set of 25 problems on many trials, we are yet to complete a comprehensive examination of these agents. More details on this analysis will be included in future updates.}. Interestingly, the agents seemed to make more mistakes (e.g., miscalculating expressions and not using each number exactly once) as the chains grew longer. Taken together, our findings highlight a fragile number sense of LLM agents despite the skills in following deterministic processes, and that the limited number sense is likely to make trial-and-error search in vast number spaces a persisting challenge for the agents.

\section{General Discussion}

Most real-world tasks (such as writing essays or code, solving mathematical problems, or driving a car) are complex. As LLM-based agents become our assistants on such tasks, understanding their complex reasoning abilities has become essential. Complex problems generally take the form of a hierarchical network of subproblems, enabling numerous possible paths from low-level actions to the high-level goals. From this perspective, reasoning can be imagined as the process of efficiently searching for appropriate paths to solutions in the hierarchy. 

In this work, we investigated LLMs' mathematical reasoning as they navigate in a "Numberland" to look for solutions. 
Specifically, we tested five leading agents on a set of 100 hand-picked problems across four categories: basic operations, advanced operations, primality checking, and a number game. The agents performed reasonably well in the first three categories, followed by a sharp drop in the number game category. These results suggest that while LLMs can effectively execute deterministic problem-solving strategies, they struggle with tasks requiring "number sense" -- an intuitive understanding of number characteristics and relationships that humans naturally possess and further develop through experience. This limitation becomes evident in the number game, which requires an informed trial-and-error search using knowledge of number factors to guide problem-solving attempts.

To verify trial-and-error search as a weakness, we presented our best performer (o1) a second set of harder 24 games, where its performance declined sharply (from 73\% to only 27\%). Examining the errors, we find that erroneously assuming the non-existence of solutions is the most common mistake, followed by breaking the rules of the game and miscalculating expressions. More recent tests on newer reasoning models indicate that the 24 game and the trial-and-error search inherent in it are persistent challenges for LLMs. Predictably, the reasoning models perform better on hard problems than the non-reasoning models. However, their improvement comes at the cost of increased computation, as they often embark on long chains of reasoning or code execution. The trade-off between reasoning capability and computational efficiency suggests the need for evaluative frameworks to consider both problem complexity and solution costs for practical applications. Moreover, despite additional computational capabilities, their performance remains poor on the harder set of 24 games. Therefore, their superiority over non-reasoning models seems to stem primarily from increases in computational resources rather than improved number senses.

Probing deeper, the types of errors made reflect a fragile number sense in LLMs. An interesting phenomenon is the breakdown of elementary mathematical skills within extended chains of reasoning. While all models included were capable of performing basic operations quite well, their skills did not always transfer to reasoning chains. They frequently miscalculated simple expressions in producing 24 and, except for the o1 model, were rarely aware of their mistakes. Moreover, they also struggled to keep count of how many times each given number was used and repeatedly broke the rule of using each number exactly once. The errors strongly suggest a fragile number sense that breaks under increased informational load during long chains of reasoning. Error patterns in other tasks also revealed some potential pitfalls, such as (1) handling uncommon exponents, especially fractional ones, (2) evaluating the primality of large numbers, (3) recognizing changes in the problem-solving environment, and (4) self-monitoring and error detection capabilities.

Generally, our results support Simon, Newell, and Shaw's insights about the hierarchical, often intractable nature of complex problems, and the need for computationally limited agents to selectively search for good-enough solutions using heuristics as optimality is beyond reach. Viewing reasoning as the ability to efficiently navigate in the hierarchy, the sharp performance drop of LLMs in puzzle games, which require flexible application of lower-level mathematical knowledge, suggests that current LLMs lack the hierarchical reasoning capabilities that humans possess in \textit{punching above their weight} in complex problems. Altogether, the findings from our relatively simple study show the merits of a divide-and-reconstruct approach for a multi-level understanding of human and AI reasoning as they grapple with complexity.



\section{Limitations and Future Directions}

While our study provides valuable insights, several key limitations should be addressed in future research. Among the vast array of LLM agents available today, we examined only a few selected ones. Even across variants of the same models, the agents can differ drastically in abilities and performance. For example, ChatGPT 4o agents could not solve some basic computations, but ChatGPT o1 agents solved them flawlessly. Future studies need to examine a larger pool of models, including different variants. Furthermore, our results need to be replicated in larger sets of computationally complex tasks to establish the apparent Achilles' heels of LLMs in trial-and-error search and, more generally, in number sense. Promising candidates for such tasks include number puzzle games like the 24 game (e.g., four fours or Sudoku), Cryptoarithmetic problems, and synthetic number problems. Alternatively, problems may be chosen based on their computational complexity (such as from the NP-Hard class of problems). 

Importantly, while we observe clear performance differences between deterministic and intuitive mathematical tasks, attributing these to specific mechanisms requires caution, as complex problems allow numerous solving strategies. For causal attribution, the paths from input information to output performance need to be traced. For this purpose, MI -- which seeks to reverse engineer LLM behavior from building blocks -- provides a promising way similar to computational cognitive modeling used to validate observations on human behavior by reconstructing it from elementary processes. Our study highlights several strengths and weakpoints in LLMs' numerical reasoning that can serve as target behavior to explain using MI. Underneath their deterministic prowess, the mistakes by agents -- such as producing non-24 outcomes or breaking 24 game rules (e.g., reusing numbers or omitting some), the lack of error awareness, and the breakdowns of reasoning chains -- suggest persisting gaps in their numerical sense. MI could unpack individual failures by identifying the building blocks (e.g., features, circuits, and motifs) involved in reasoning and examining how these internal representations drift during extended reasoning to offer insights beyond traditional evaluation methods. Such comprehensive examinations help integrate different explanations of complex reasoning for a unified understanding.


\section{Summary and Conclusions}

In this work, we examined the numerical reasoning abilities of LLM-based agents in Numberland. The results of the experiments demonstrated both the strengths and limitations of numerical reasoning in Large Language Models (LLMs). While LLMs performed well on tasks involving deterministic operations such as basic arithmetic, advanced calculations, and prime number checks, their accuracy dropped significantly when tested on more complex tasks like the 24 game, which requires trial-and-error reasoning. These findings highlight the disparity between human and AI performance, particularly when handling uncertainty and multi-step problem-solving without prior fine-tuning.

More generally, our study highlights the merits of conducting targeted experiments such as our "Numberland" to systematically evaluate and monitor LLMs' numerical prowess in navigating complex spaces. In future studies, researchers may further elucidate the challenges LLM agents face in numerical reasoning by specifying their roots in the building blocks. Together, these explorations may pave the way for improved AI safety by showing ways to improve LLM reasoning, as well as by drawing clearer lines in the sand between AI and human reasoning in complex real-world problems.

\section{Acknowledgments}
I am sincerely grateful to Michael J. Schoelles and Jeff Shrager for sharing their vast knowledge of mathematical reasoning and providing invaluable guidance in this research. I also thank Ummay H. Tabassum, my dear wife, for her insights on LLM reasoning. This study was conducted using personal time and resources. LLM-based AI agents (Grok, Claude, and Copilot) were used to assist in draft review, formatting, and grammar checks. The data generated in this work can be found here: \url{https://osf.io/fncyu/}. The problem sets of Numberland are included in Appendices \ref{sec:prompts_used} and \ref{sec:prompts_used_followup}. We highly encourage the readers to test their LLM assistants on these problems, especially the games of 24.

\printbibliography

\newpage
\appendix

\section{Prompts Used in Main Experiment}\label{sec:prompts_used}

Here, we provide the experimental script of prompts used to communicate with the LLMs.\\

\noindent\underline{Prompt 1}: Hi. I will give you some sets of mathematical problems. Solve them as best as you can. For any fractions, provide 3 digits after the decimal point. You are welcome to provide the reasonings, but please remember to summarize all answers in a list following the order of the problems presented. Do you understand?\\

(After an affirmative response from the agent)\\

\noindent\underline{Prompt 2}: Here is the first set of problems. Simply calculate these values.

\begin{enumerate}
	\item $1+2-3*4/5$
	\item $2+2+4-2$
	\item $(3+4*3)/5$
	\item $1/13$
	\item $13*7/(169*25)$
	\item $(13+4)(13-4)$
	\item $1/1.2+1/6$
	\item $(1/13-1/11)*999999$
	\item $87/29+2$
	\item $49\pi$
	\item $\pi/6$
	\item $e+\pi$
	\item $79/13$
	\item $1+2+3+4+5+6+7$
	\item $1-2+3-4+5-6+7-8$
	\item $1-1+1-1+1$
	\item $55 -1-2-3-4-5-6-7-8-9$
	\item $1+1/2+1/3+1/4+1/5+1/6+1/7$
	\item $1/10+1/100+1/1000+1/10000$
	\item $11*101/9999$
	\item $1/(11*101*73*137)$
	\item $(11*13*17*19)/(22*39*68*95)$
	\item $1/(3/10)-20/6$
	\item $1/(2*12-3*(73-69))$
	\item $1/(2*12-2*3*(73-69))$
\end{enumerate}
\noindent\underline{Prompt 3}: Okay. Now, calculate these values.

\begin{enumerate}
	\item $0^0$
	\item $1^0$
	\item $2^3$
	\item $2^{-5}$
	\item $400^{-.5}$
	\item $400^{0.23}$
	\item $237^{3.7}$
	\item $500^{2/7}$
	\item $300^{3.1416}$
	\item $700^{-1/7}$
	\item $71^{0.26}$
	\item $4900\pi^2$
	\item $(\pi +2)(\pi-2)-\pi^2$
	\item $\pi/10 + log(e^{\pi/5})$
	\item $71^{0.26e^{\pi i}}$
	\item $71^{0.26e^{2\pi i}}$
	\item $log_2 (2^{0.23})$
	\item $log_{23} (0.309)$
	\item $log_\pi((1+i)^2-2i)$
	\item $log_8 (512)$
	\item $log_5(1/625)$
	\item $log_5(1/625)$
	\item $log_2 (2^{20000000000000000000000000000000000000})$
	\item $log_2 (2^{20000000000000000000000000000000000000i})$
	\item $log_{13}(999999/(7*27*37*11))$
\end{enumerate}

\noindent\underline{Prompt 4}: Now, check if these numbers are prime or not.

\begin{enumerate}
	\item $5$
	\item $233$
	\item $349$
	\item $361$
	\item $367$
	\item $499$
	\item $71993$
	\item $1282529$
	\item $3326489$
	\item $514229$
	\item $91193$
	\item $99194853094755497$
	\item $1066340417491710595814572169$
	\item $263130836933693530167218012159999999$
	\item $359334085968622831041960188598043661065388726959079837$
	\item $2^7-1$
	\item $2^{13}-1$
	\item $2^{17}-1$
	\item $2^{19}-1$
	\item $2^{31}-1$
	\item $2^{11}-1$
	\item $2^{23}-1$
	\item $2^{29}-1$
	\item $2^{37}-1$
	\item $2^{27}-1$
\end{enumerate}

\noindent\underline{Prompt 5}: Great! The next set is a bit different. Are you familiar with the 24 game and its rules? If you are, I will give you 25 games to play (i.e., 25 sets of 4 digits to make 24 with).\\

(After an affirmative response from the agent and verifying that they know the rules)\\

\noindent\underline{Prompt 6}: Perfect. You are welcome to provide the reasonings, but please remember to summarize all answers in a list following the order of the problems presented. Here is a set of 25 problems.

\begin{enumerate}
	\item $1,1,1,24$
	\item $1,1,2,12$
	\item $1,2,3,4$
	\item $1,1,4,6$
	\item $1,1,3,8$
	\item $0,15,39, 81$
	\item $3,3,4,5$
	\item $2,2,2,8$
	\item $6,6,8,12$
	\item $1,3,3,3$
	\item $4,5,9,13$
	\item $1,1,1,12$
	\item $5,9,12,13$
	\item $6,6,8,8$
	\item $4,6,7,9$
	\item $7,8,9,12$
	\item $2,4,6,6$
	\item $6,7,8,12$
	\item $1,2,3, 13$
	\item $3,9,12,13$
	\item $1,1,6,12$
	\item $1,5,6,12$
	\item $1,11,12,13$
	\item $2,2,8,8$
	\item $2,8,8,9$
\end{enumerate}

\newpage
\section{Prompts used in Follow-up Experiments}\label{sec:prompts_used_followup}

\underline{Only Prompt}: Are you familiar with the 24 game and its rules? If you are, I have 25 games for you to play (i.e., 25 sets of 4 numbers to make 24 with). First, state the rules briefly, and then start playing!

\begin{enumerate}
	\item $7,10,12,13$
	\item $4,4,8,9$
	\item $2,8,10,12$
	\item $2,3,9,12$
	\item $5,9,12,13$
	\item $3, 5, 9, 10$
	\item $2,5,7,8$
	\item $1,4,9,13$
	\item $5,9,10,11$
	\item $2,3,8,13$
	\item $5, 10, 10, 13$
	\item $2,3,8,13$
	\item $3,5,8,13$
	\item $3,5,9,10$
	\item $4,5,6,12$
	\item $2,2,3,5$
	\item $2,4,7,8$
	\item $3,4,5,11$
	\item $4,4,7,7$
	\item $2,7,7,10$
	\item $2,5,10,12$
	\item $3,9,9,11$
	\item $2,9,13,13$
	\item $2,3,9,12$
	\item $5,7,9,10$
\end{enumerate}






\end{document}